\documentclass{article}

\usepackage[preprint]{neurips_2021}

\usepackage[utf8]{inputenc} 
\usepackage[T1]{fontenc}    
\usepackage{hyperref}       
\usepackage{url}            
\usepackage{booktabs}       
\usepackage{bbm}
\usepackage{amsfonts}       
\usepackage{nicefrac}       
\usepackage{microtype}      
\usepackage{xcolor}         
\usepackage[pdftex]{graphicx}

\title{Multiclass ASMA vs Targeted PGD Attack in Image Segmentation}

\author{
  Johnson Vo \\
  University of Toronto\\
  \texttt{johnson.vo@mail.utoronto.ca} \\
  \And 
  Jiabao Xie \\
  University of Toronto\\
  \texttt{jimbo.xie@mail.utoronto.ca} \\
  \And
  Sahil Patel \\
  University of Toronto\\
  \texttt{sas.patel@mail.utoronto.ca} \\
}

\begin{document}

\maketitle

\begin{abstract}
Deep learning networks have demonstrated high performance in a large variety of applications, such as image classification, speech recognition, and natural language processing. However, there exists a major vulnerability exploited by the use of adversarial attacks. An adversarial attack imputes  images by altering the input image very slightly, making it nearly undetectable to the naked eye, but results in a very different classification by the network. This paper explores the projected gradient descent (PGD) attack and the Adaptive Mask Segmentation Attack (ASMA) on the image segmentation DeepLabV3 model using two types of architectures: MobileNetV3 and ResNet50, It was found that PGD was very consistent in changing the segmentation to be its target while the generalization of ASMA to a multiclass target was not as effective. The existence of such attack however puts all of image classification deep learning networks in danger of exploitation.
\end{abstract}

\section{Introduction}
 Despite the prevalence of adversarial attacks in modern research across image classification, its implementation on image segmentation models have not been fully explored. Image segmentation is similar to standard classification, however instead it identifies which specific pixels in an image belong to a specific class, segmented out of the image [1]. Inspecting adversarial attacks within image segmentation is incredibly important as it could increase the robustness of future segmentation models to make them more reliable. For instance, adversarial attacks against an image segmentation network in the medical field could result in a poor model failing to identify a potential tumor.

To investigate adversarial attacks in image segmentation, we look into the projected gradient descent (PGD) attack and (ASMA) on the DeepLabV3 [4] model using two types of architectures: MobileNetV3 [3] and ResNet50. These two backbones were chosen as a result of their stark difference in purpose. While the MobileNetV3 architecture emphasizes computational efficiency, ResNet50 tries to maximize accuracy by using a much larger model. Trained models on the Pascal Dataset are used to compare the attack methodologies, on the amount of perturbation on a given image they impose. 
\section{Targeted Projected Gradient Descent Attack}
This adversarial attack aims to change the result of a segmentation prediction to a proposed target. Unlike other attacks, this targeted attack is powerful because achieving a specific and important classification can grant access to otherwise forbidden material, such as face recognition networks. This attack requires access to the original model and thus is classified as a white-box attack; a type of attack where adversarial examples are trained specifically for each model. With most adversarial attacks, PGD imputes the original image by adding invisible noise through the use of a perturbation. Projected gradient descent (PGD) generates this by minimizing the cross-entropy loss between the segmentation mask created by the perturbed image (original image + perturbation), and the segmentation mask of a completely separate image (the target mask). We can rigorously define the goal of our gradient descent to be to find a perturbation $P$ such that:
$$P = argmin_{|P| < \varepsilon}(\mathcal{L}(g(X+P), Y_t))$$
where we have that $\mathcal{L}()$ represents the cross-entropy loss between the output segmentation mask of a neural network $g(X+P)$ where $X$ is the original input image, $P$ is the perturbation, and $Y_t$ is the target segmentation mask. The image input $X$ is of size $B \times C \times H \times W$ where $B$ is the batch size, $C$ representing the number of channels, and $H$, and $W$ for height and width of the image. This methodology restricts the perturbed image by pixel values at most $\varepsilon$ away from the image, to ensure the perturbed image wouldn't look too visually different.

Gradient descent for PGD is conducted similarly to normal gradient descent, guided with a learning parameter $\alpha$. We calculate the perturbation at step $n$, such that:
$$P_n = P_{n-1} - \alpha \cdot \nabla P_{n-1}$$
Where we try to minimize the cross entropy loss between $g(X + P_n)$ and $Y_t$. Thus to determine how to create effective image segmentation attacks using PGD, we will compare it to the results of the ASMA attack described below.
\section{Adaptive Segmentation Mask Attack (ASMA)}

As proposed in Ozbulak et al's work on biomedical image segmentation [2], the Adaptive Segmentation Mask Attack is a targeted adversarial white box attack primarily concerned with precision. Its attack, performed on a binary segmentation model, was powerful in misclassifying images very consistently. We generalize this attack to the multiclass segmentation case. ASMA is somewhat similar to PGD in that its goal is to train adversarial examples to have a similar segmentation mask to a non-related targeted segmentation mask. However, instead of restricting the perturbation to be less than some $\varepsilon$, we use gradient descent to minimize our perturbation, $||P||_2$ and minimize the the cross entropy loss between $g(X+P)$ and $Y_t$.
At iteration $n$ in gradient descent, we define our perturbed image (with the input image $X$ of same shape as in PGD) to be $X_{n+1} = X_n + \alpha P_n$, where $\alpha$ is a learning rate hyperparameter. We then define $P_n$ such that we increase the probability of segmenting pixels to match the targeted segmentation mask (and avoid optimizing pixels that are already correct in the mask). We do this by defining the gradient of $P_n$ such that:

\begin{equation}
    P_n = \sum_{c=0}^{M-1} \nabla_x(g(X_n)_c \odot \mathbbm{1}_{Y_t=c} \odot \mathbbm{1}_{argmax_M(g(x_n))\neq c})
\end{equation}

Where $\odot$ is the element wise product, and $M$ is the number of segmentation classes (21 in our dataset). Constructing the perturbation this way ensures that the gradient is only retrieved from pixels where the the segmentation isn't already the target classification, increasing precision. Lastly, the ASMA methodology iteratively defines $\alpha$ dynamically to take into account the progress of masking. Using a perturbation multiplier, at iteration $n$, we have that:
\begin{equation}
    \alpha_n = \beta \cdot IoU(Y_t, g(X_n)) + \tau
\end{equation}

Where $\beta$ and $\tau$ are hyperparameters, and the intersection over union (IoU) is calculated and used to determine the percentage of the same classified pixels between two segmentation masks.
\section{Models}
Given that the aim of this paper is to compare these attacks on image segmentation models, we use one of the strongest available architects: DeepLabV3 [4]. This architecture uses multiple atrous convolution layers that are layered on top of each other, and is called the atrous spatial pyramid pooling. This has been shown to be effective in detecting spatial patterns regardless of scale. The DeepLabV3 architecture also uses Conditional Random Fields to sharpen boundaries by incoporating surrounding pixels within the classification of a single pixel. With regards to implementation of this model, it has been done with a MobileNetV3 and ResNet50 backbone that each have their own advantages.
\subsection{MobileNetV3 Architecture}
The purpose of the MobileNet backbone is to implement a given architecture such that it can be used on less powerful and CPU based mobile devices, like a phone. To reduce the size of the model, depthwise separable convolutions are used. These convolutions can be considered  a depthwise convolution followed by a pointwise convolution. Proceeding in this manner drastically reduces the total number of model parameters. However, it may lack the capability to learn more complex patterns in a network given the limitation of resources.
\subsection{ResNet50 Architecture}
Quite opposite to the MobileNetV3 architecture, ResNet50 aims to make the most accurate results. As a result of not being meant for a mobile device, ResNet50 has 50 hidden layers and is designed using basic convolution layers along with batch normalization and ReLU layers. Furthermore, ResNet50 makes use of skip-connections in an attempt to avoid the vanishing gradient problem.
\section{Experiment}

We use pretrained Pytorch implementations of DeepLabV3 using a MobileNetV3 or ResNet50 architecture. These models were trained on the COCO 2017 dataset which contains over 100 thousand different images, with 21 different segmentation classes to be predicted. Next, a small batch of the same images are trained to be adversarial examples. It is noted that convergence for ASMA is reached with $\tau = 1e-5, \beta= 3e-6$ for the MobileNetV3 architecture, while $\tau = 1e-5, \beta= 1e-4$ for the ResNet50 architecture, while $\varepsilon = 0.1$ and $\alpha=100$ is used based on previous implementations for the PGD attacks. Various diagnostic values were calculated to determine the performance of the attack image once the adversarial examples were trained. IoU was calculated to determine how close the adversarial mask was to the original mask of the image. Pixel accuracy was calculated for a similar purpose. Lastly, the L2 norm between the original images and their adversarial counterparts was calculated to quantify the amount of alteration to the original image that was made.

\subsection{Training}

The total loss in our implementation of PDG, with inspiration of existing implementation [5] is the loss between every image's segmentation and the targeted image segmentation. As the attack iterates, the loss decreases as the perturbed image segmentation becomes more closely resembled to the targeted segmentation. The loss therefore reaches quite low and the perturbations are effectively working. An figure of the loss is found in Figure 2 of the appendix. 

\begin{figure}[h]
  \centering
   \frame{\includegraphics[width=1\linewidth]{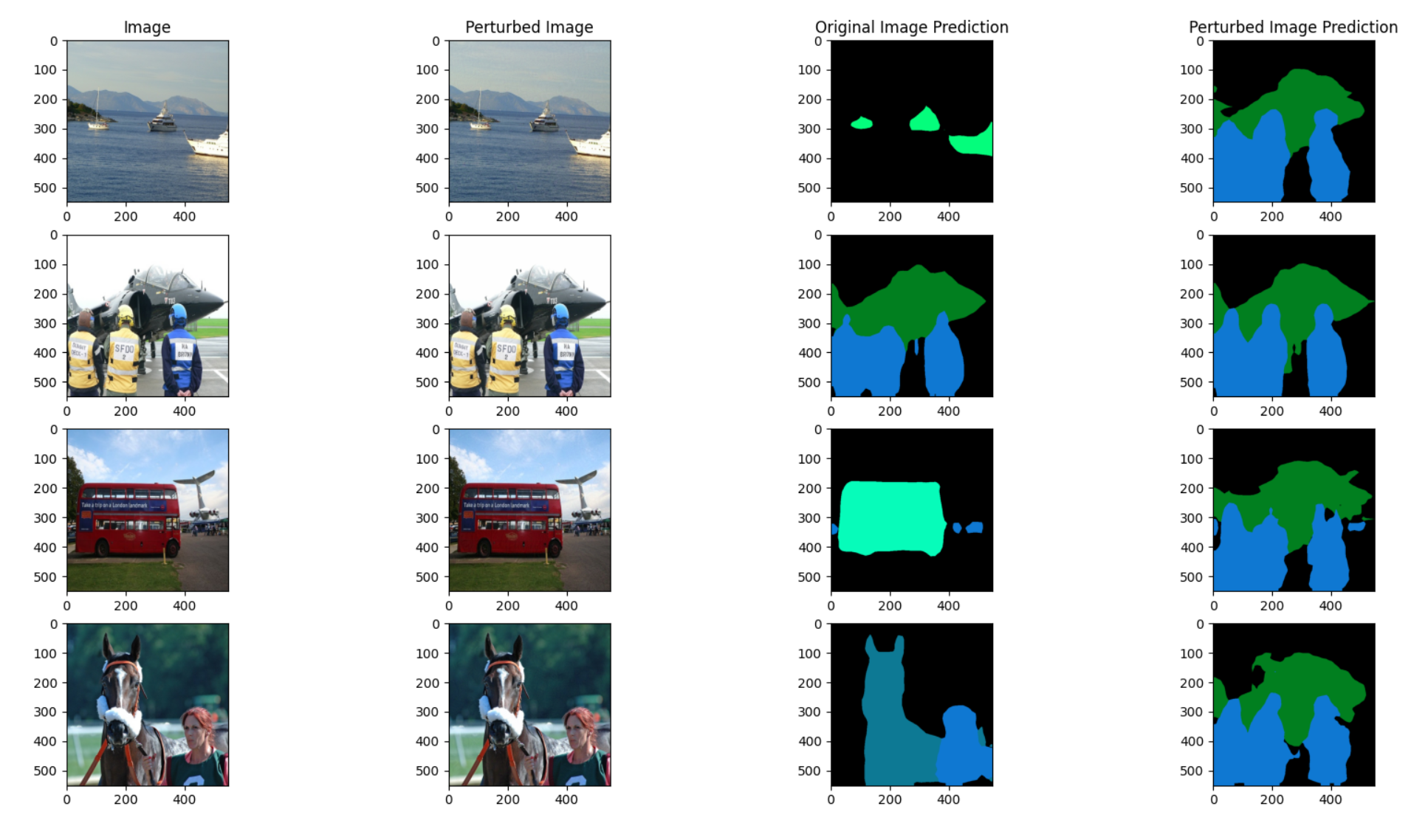}}
  \caption{MobileNet Output}
\end{figure}

We can visualize the effect of the PGD attack on the image segmentation in Figure 1 performed on the MobileNet model. In the first column is the original input images. The second column is the perturbed image as the result of the adversarial attacks. The third column is the image segmentation and prediction from the model with the original image as input. Similarly the third column is the model prediction for the perturbed images. In every case, the target image (and thus the target mask) are the images found in the second row, and qualitatively each perturbed image segmentation is very recognizable as the target image's segmentation. The shape of the plane and the three humans' segmentation are visible from the perturbed image.

\section{Results}

\begin{table}[h]
\caption{Segmentation Attack Diagnostic Values}
\begin{tabular}{llll}
\hline
Architectureand Attack Type & IoU (\%) & Pixel Accuracy (\%) & Perturbation L2 Norm \\ \hline
ResNet ASMA & 36.93 & 55.82 & 3534895.25 \\ \hline
ResNet PGD Attack & 15.75 & 55.53 & 5171.84 \\ \hline
MobileNet ASMA & 42.31 & 58.85 & 7650.44 \\ \hline
MobileNet PGD Attack & 14.88 & 53.39 & 5020.88 \\ \hline
\end{tabular}
\end{table}

As seen in Table 1, the diagnostic values heavily dependent on whether the adversarial image was generated using ASMA or the PGD attack method. Notice that pixel accuracy is relatively similar across all models and corresponding attacks. However, for ASMA the IoU is way higher, implying that the final segmentation mask still has some relation to the original segmentation mask, suggesting that there is less change in the segmentation mask. Lastly, the perturbation L2 norm from ASMA is significantly higher, hence the adversarial image has a greater visual difference than what the difference between the original image and the PGD generated image. See the appendix figure 3-6 for its illustration. Not only can slight differences in the original and perturbed image be seen clearly in the ASMA generated plots, but failure to become the target mask is also visible. 

\section{Discussion}
After expanding out the application of ASMA from the binary case, into a multi-class segmentation scheme, the generated adversarial examples seemingly are  worse than attack images that were generated using the extremely common PGD attack approach. As a result, it is reasonable to believe that given how the ASMA approach is not restricted in its descent range, that it performs better among a narrower spread of image types. For example, the original proposition of ASMA was over a dataset with similar types of images (ophthalmological in nature). By equation (1), the perturbation of the image is sourced on the gradient of class $c$ of pixels that are different from the targeted mask. Due to the large number of segmentation classes, this ends up with a very sparse perturbation as opposed to other attacks, which weakens the effect of the attack. On models with similar inputs and smaller number of classes, the ASMA attack is suspected to perform better. To improve results, performing the attack on more similar looking inputs to generalize to the multiclass case is a potential step. In future research into how ASMA or PGD can be used in different segmentation problems should be conducted as to define the best choice of adversarial examples.

\newpage
\section{References}
[1] Image segmentation : Tensorflow Core. TensorFlow. (2022, January 26). Retrieved April 20, 2022, from https://www.tensorflow.org/tutorials/images/segmentation 

[2] Ozbulak, U., Van Messem, A., \& De Neve, W. (2019). Impact of Adversarial Examples on Deep Learning Models for Biomedical Image Segmentation. arXiv. https://doi.org/10.48550/ARXIV.1907.13124

[3]Massa, V. V. and F. (2022). Pytorch. PyTorch. Retrieved April 20, 2022, from https://pytorch.org/blog/torchvision-mobilenet-v3-implementation

[4] Chen, L.-C., Papandreou, G., Schroff, F., \& Adam, H. (2017). Rethinking Atrous Convolution for Semantic Image Segmentation. arXiv. https://doi.org/10.48550/ARXIV.1706.05587

[5] Chada, S. (2021). Adversarial examples on Semantic Segmentation Networks. GitHub. Retrieved April 20, 2022, from https://github.com/Swaraj-72/Adversarial-Attacks-and-Defenses-on-Semantic-Segmentation-Networks

\newpage
\appendix

\section{Appendix}
\subsection{Contributions}
\begin{itemize}
    \item Sahil Patel was primarily responsible for writing and running the necessary created programs
    \item Johnson Vo contributed in coding parts of PDG, ASMA, and writing about the results of ASMA.
    \item Jiabao Xie contributed in coding parts of PDG, experimenting with ASMA and writing the results on PDG.
\end{itemize}

\subsection{Figures}

\begin{figure}[h]
  \centering
   \frame{\includegraphics[width=200pt]{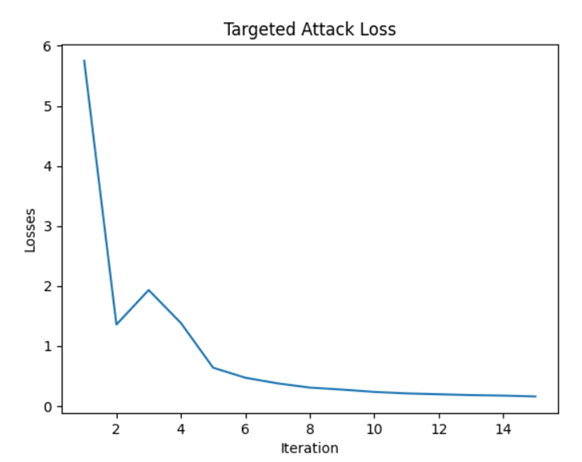}}
  \caption{MobileNet Loss on PDG}

\end{figure}

\begin{figure}[h]
  \centering
   \frame{\includegraphics[width=1\linewidth]{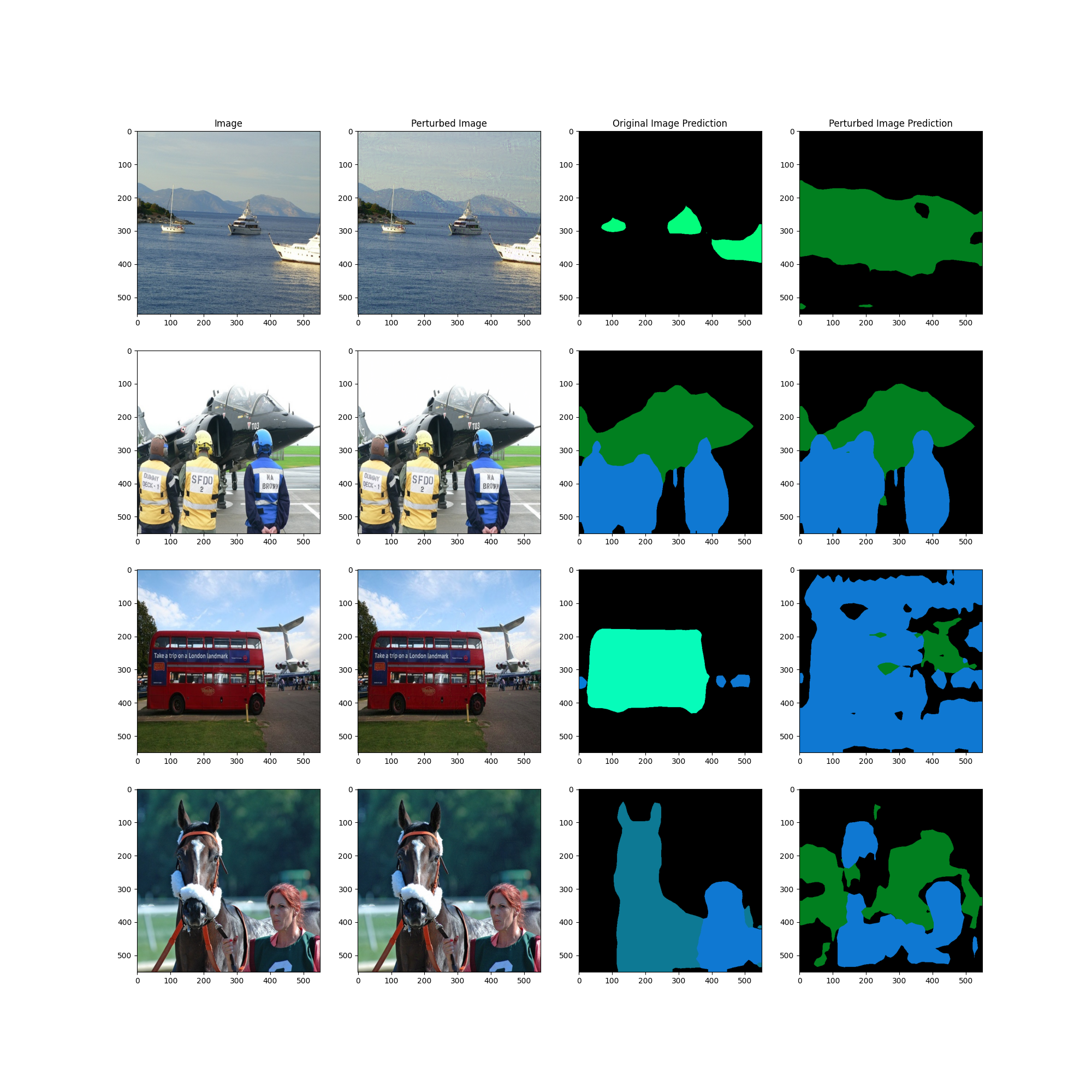}}
  \caption{MobileNet Output after ASMA}
\end{figure}

\begin{figure}[h]
  \centering
   \frame{\includegraphics[width=1\linewidth]{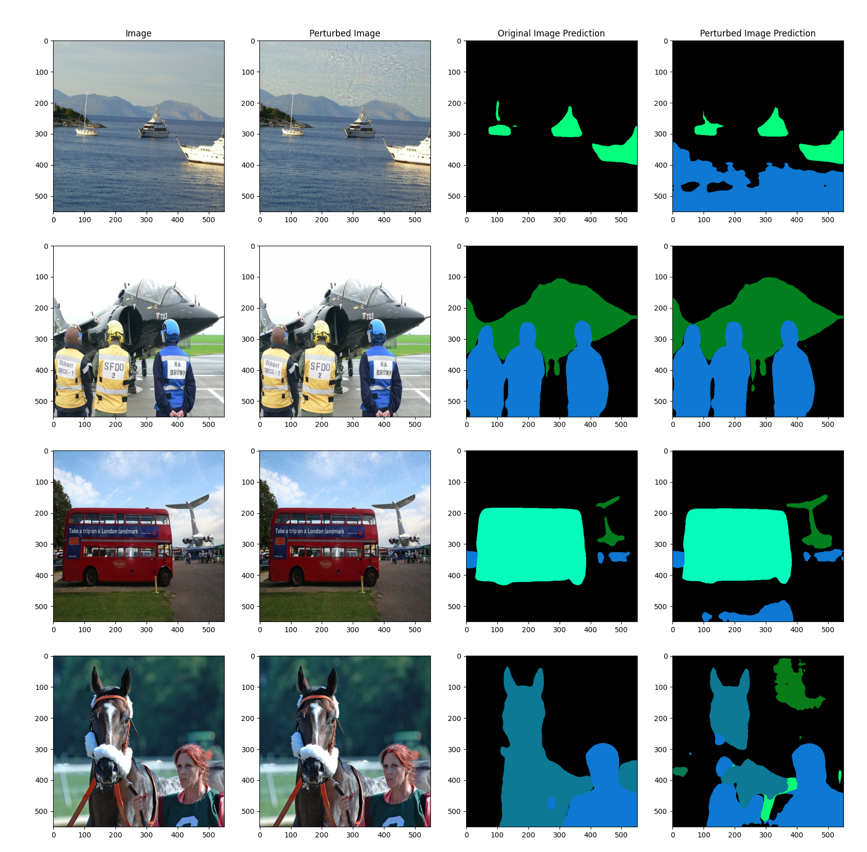}}
  \caption{ResNet Output after ASMA}
\end{figure}

\begin{figure}[h]
  \centering
   \frame{\includegraphics[width=1\linewidth]{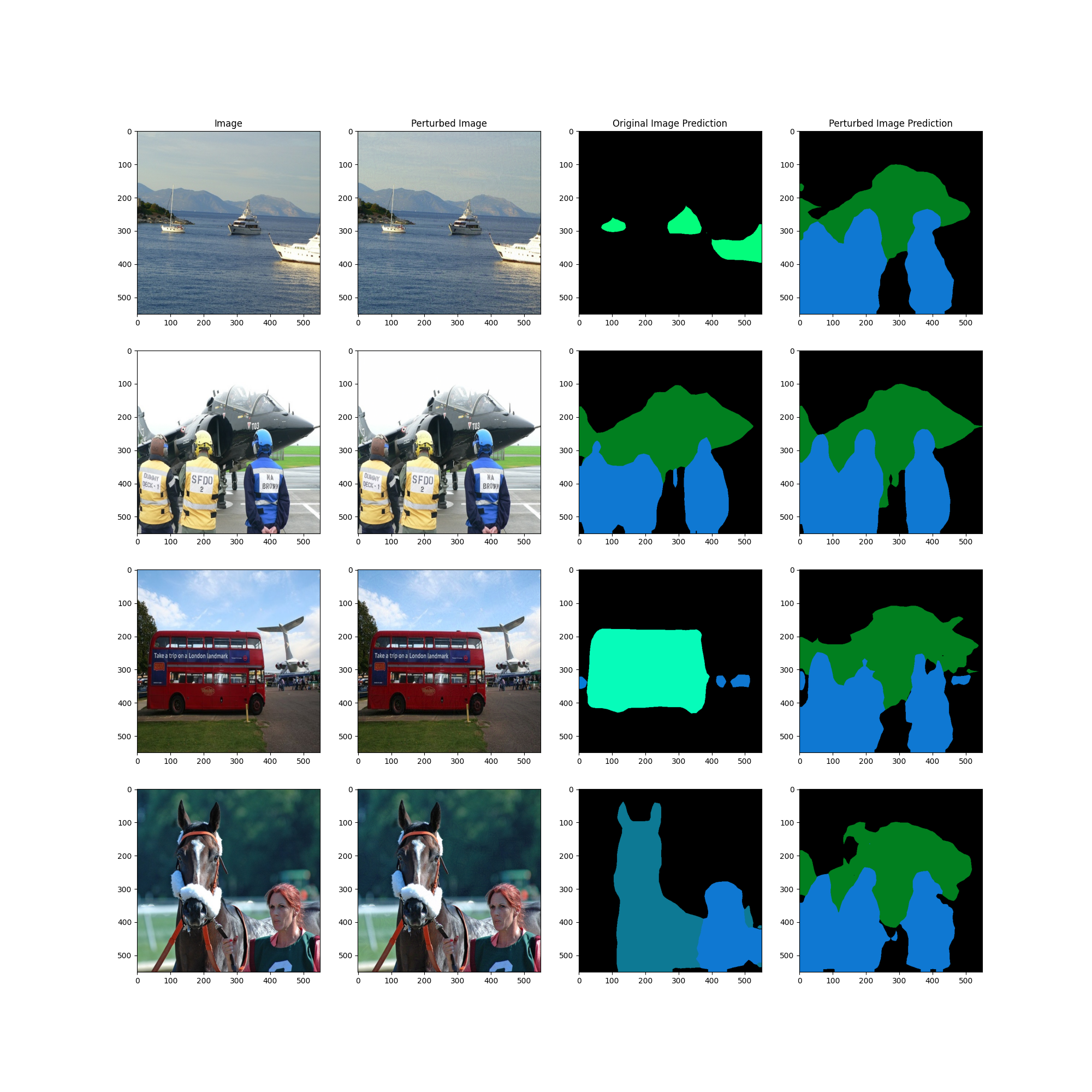}}
  \caption{MobileNet Output after PGD Attack}
\end{figure}

\begin{figure}[h]
  \centering
   \frame{\includegraphics[width=1\linewidth]{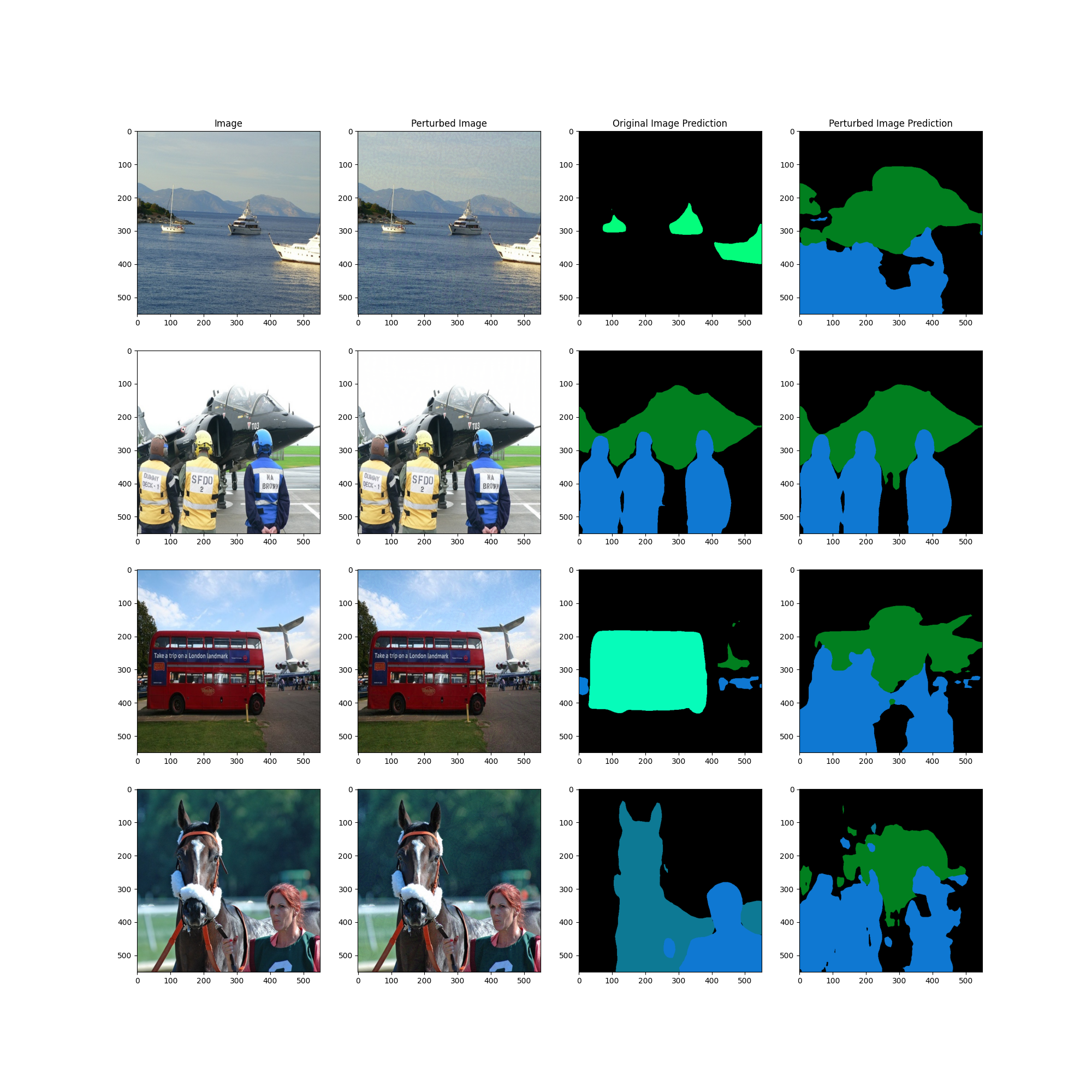}}
  \caption{ResNet Output after PGD Attack}
\end{figure}

\end{document}